\documentclass{article}

\usepackage[preprint]{neurips_2024}

\usepackage[utf8]{inputenc} % allow utf-8 input

\usepackage{lineno,subcaption}
\usepackage[T1]{fontenc}
\usepackage{graphicx}
\usepackage{booktabs}
\usepackage[misc]{ifsym}

\usepackage{mwe}
\usepackage{url}
\usepackage{amsfonts, amsmath, amssymb}
\usepackage{mathtools}
\usepackage{nicefrac}       % compact symbols for 1/2, etc.
\usepackage{microtype}      % microtypography
\usepackage{xcolor}  
\usepackage{booktabs}
\usepackage{float}
\usepackage{hyperref}

\usepackage{listings}
\usepackage{dirtree}

\definecolor{codegreen}{rgb}{0,0.6,0}
\definecolor{codegray}{rgb}{0.5,0.5,0.5}
\definecolor{codepurple}{rgb}{0.58,0,0.82}
\definecolor{backcolour}{rgb}{0.95,0.95,0.92}

\lstdefinestyle{mystyle}{
    backgroundcolor=\color{backcolour},   
    commentstyle=\color{codegreen},
    keywordstyle=\color{magenta},
    numberstyle=\tiny\color{codegray},
    stringstyle=\color{codepurple},
    basicstyle=\ttfamily\footnotesize,
    breakatwhitespace=false,         
    breaklines=true,                 
    captionpos=b,                    
    keepspaces=true,                                 
    showspaces=false,                
    showstringspaces=false,
    showtabs=false,                  
    tabsize=2
}

\title{Benchmarking the Influence of Pre-training on Explanation Performance in MR Image Classification}

\author{%
  Marta Oliveira \\
  Physikalisch-Technische Bundesanstalt\\
  Abbestr. 2–12, 10587 Berlin, Germany\\
  \And
  Rick Wilming \\
  Technische Universität Berlin\\
  Str. des 17. Juni 135, 10623 Berlin, Germany\\
  \And
  Benedict Clark \\
    Physikalisch-Technische Bundesanstalt\\
  Abbestr. 2–12, 10587 Berlin, Germany\\
  \And
  Céline Budding, Fabian Eitel, Kerstin Ritter \\
  Charité – Universitätsmedizin Berlin\\
  Charitéplatz 1, 10117 Berlin, Germany \\
  \And
  Stefan Haufe \\
    Physikalisch-Technische Bundesanstalt\\
  Abbestr. 2–12, 10587 Berlin, Germany\\
  Technische Universität Berlin\\
  Str. des 17. Juni 135, 10623 Berlin, Germany\\
  Charité – Universitätsmedizin Berlin\\
  Charitéplatz 1, 10117 Berlin, Germany \\
  \texttt{haufe@tu-berlin.de}
  \
}

\begin{document}

\maketitle

\begin{abstract}
Convolutional Neural Networks (CNNs) are frequently and successfully used in medical prediction tasks. They are often used in combination with transfer learning, leading to improved performance when training data for the task are scarce. The resulting models are highly complex and typically do not provide any insight into their predictive mechanisms, motivating the field of `explainable' artificial intelligence (XAI). 
However, previous studies have rarely quantitatively evaluated the `explanation performance' of XAI methods against ground-truth data, and transfer learning and its influence on objective measures of explanation performance has not been investigated. 
Here, we propose a benchmark dataset that allows for quantifying explanation performance in a realistic magnetic resonance imaging (MRI) classification task.
We employ this benchmark to understand the influence of transfer learning on the quality of explanations. 
Experimental results show that popular XAI methods applied to the same underlying model differ vastly in performance, even when considering only correctly classified examples.
We further observe that explanation performance strongly depends on the task used for pre-training and the number of CNN layers pre-trained. 
These results hold after correcting for a substantial correlation between explanation and classification performance.

\end{abstract}

\section{Introduction}
\begin{figure}
    \centering
    \includegraphics[width=0.9\linewidth]{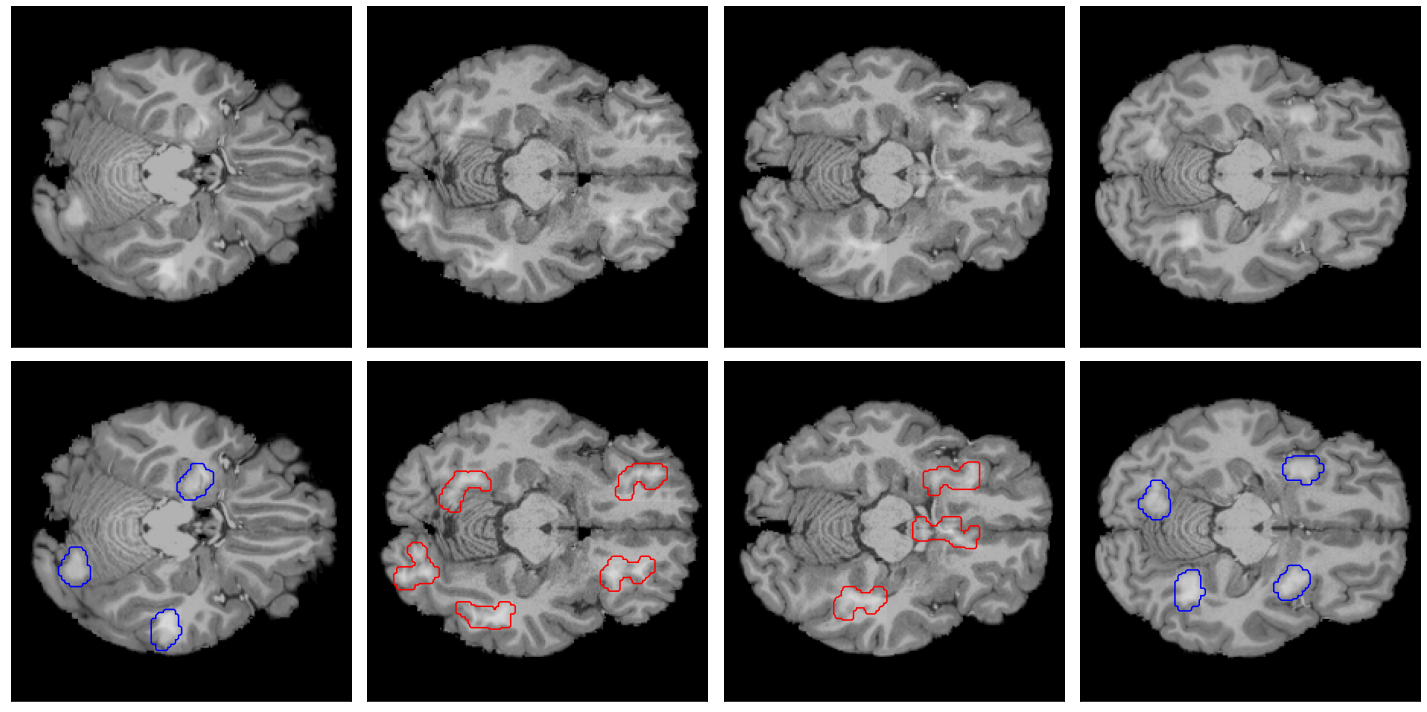}
    \caption{Example of images of the dataset created. The top row consists of axial MRI slices from the Human Connectome Project \citep[HCP,][]{hcp} healthy brain dataset, with artificial lesions added. The bottom row consists of the top row images, but with the position of these lesions contoured in blue, forming the ground-truth for an explanation.}
    \label{fig:example_images} 
\end{figure}

Following AlexNet's \citep{krizhevskyImageNetClassificationDeep2012} victory in the ImageNet competition, 
CNNs developed to become the deep neural network (DNN) architecture of choice for any image-based prediction tasks. Apart from their ingenious design, the success of CNNs was made possible by ever-growing supplies of data and computational resources. 
However, sufficient labelled data to train complex CNNs are not widely available for every prediction task. This is especially true for medical imaging data, which is cumbersome to acquire and underlies strict data protection regulations. To address this bottleneck, Transfer Learning (TL) techniques are frequently employed \citep{TL_med}. In the context of DNNs, TL strategies often consist of two steps. First, a surrogate model is trained on a different prediction tasks, for which ample training data are available. This is called pre-training. And, second, the resulting model is adapted to the prediction task of interest, where only parts of the model's parameters are updated, while other parameters are kept untouched \citep{TL_main}.
This is called re-training or fine-tuning and requires smaller amounts of labelled data than training a network from scratch, leading to a less computationally expensive process. TL is, therefore, frequently employed for prediction tasks in medical imaging, where it is believed that TL techniques improves the generalisation by identifying common features between the two tasks \citep{1}. \citet{example-finetuning-ultrasound} use a DNN, trained with the ImageNet dataset \citep{imagenet}, to classify ultrasound images into eleven categories, achieving better results than human radiologists. Another example is the reconstruction of Magnetic Resonance Imaging (MRI) data with models trained on an image corpus that was augmented with ImageNet data \citep{example-finetuning-MRI}. The resulting model outperformed conventional reconstruction techniques.
However, it also has been argued \citep{train_with_MRI} that the use of pre-trained models may not be adequate for the medical field. The main argument being that structures in medical images are very different from those observed in natural images. Hence, feature representations learned during pre-training may not be useful for solving clinical tasks. 

Despite the success of DNN models, their intrinsic structure makes them hard to interpret. This challenges their real-world applicability in high-stake fields such as medicine. Although many practices in medicine are still not purely evidence-based, the risk posed by faulty algorithms is exponentially higher than that of doctor–patient interactions \citep{exponential_risk}. Thus, it has been recognised that the working principles of complex learning algorithms need to be made transparent if such algorithms are to be used on critical human data. The General Data Protection Regulation of the European Union (GDPR, Article 15), for example, states that patients have the right to receive meaningful information about how decisions are achieved based on their data, including decisions made based on artificial intelligence algorithms, such as DNNs \citep{GDPR0218}.

 \textcolor{black}{The field of `explainable artificial intelligence' (XAI) arose to address this need. A popular class of XAI methods seek to deliver so-called local post-hoc explanations, which are derived from a trained model's output on a test input. 
These methods can be either specific to a particular architecture or type of ML model or model-agnostic, where explanations can be produced for a large variety of model architectures.
The outcome of such methods is often a so-called heat map, which assigns an `importance' score to each input feature.
However, despite the popularity of XAI methods, their theoretical underpinnings are far from established. Most importantly, there is no agreed upon definition of what explainability means or what XAI methods are supposed to deliver \citep{lim2}. Consequently, little quantitative empirical validation of XAI methods exists \citep{lim3}. This limits the utility of XAI methods for quality control purposes in critical domains such as medicine. }

 \textcolor{black}{To date, most quantitative evaluations of XAI methods focus on secondary quality aspects such as robustness or uncertainty of explanations but spare out the fundamental issue of explanation correctness. To define a notion of correctness, it is necessary to devise working definitions of what constitutes a desirable explanation for a given input datum. Such a definition would allow one to measure the explanation performance of XAI methods using objective metrics. Synthetic data whose data-generating process is known by construction provide such a ground truth.}
 
 \textcolor{black}{In this work, we focus on the problem of classifying MR images of the human brain. We devise synthetic ground-truth data for this problem, thereby addressing the current lack of validation of model explanations in this context. Precisely, we overlay real MR images with artificial lesions of two different types, where the type of lesion defines class membership. Lesions are realistically designed to resemble white matter hyperintensities (WMH), which are important biomarkers of the aging brain and ageing-related neurodegenerative disorders \cite{WMH2,WMH1}. 
As the positions of the class-discriminative lesions are fully known by construction this provide a ground-truth for model explanations. 
We provide an open code and data framework for generating MRI slices with different types of lesions and respective ground-truths\footnote{ \textcolor{black}{Please find all code here: \url{https://github.com/Marta54/Pretrain_XAI_gt}, and all  benchmark data here: \url{https://www.doi.org/10.17605/OSF.IO/XNWAJ}}}.}

 \textcolor{black}{In the second part of this work, we show the benchmark's utility by investigating both classification and explanation performance as a function of model pre-training. Concretely, we benchmark common XAI methods against each other and compare the explanation performance of models pre-trained using either within-domain data (using different MRI classification tasks) or out-of-domain data \citep[using natural images from the ImageNet classification challenge,][]{imagenet} as well as models that have been retrained on the task of interest to varying degrees.}

\section{Related Work}
 \textcolor{black}{A number of recent works in the field of XAI have moved towards objective validation of XAI approaches using synthetic data. 
}
\textcolor{black}{\citet{kimInterpretabilityFeatureAttribution2018} propose to validate XAI via surrogate ground-truth information employing so called concept activation vectors (TCAV), which are accessible with synthetic data.
\citet{yangBenchmarkingAttributionMethods2019} propose a notion of relative feature importance to develop a metric to quantitively assess methods such as TCAV.
Known data generating processes are also increasingly being utilized to provide ground-truth information for model explanations \citep{ismailInputCellAttentionReduces2019, ismailBenchmarkingDeepLearning2020, tjoaQuantifyingExplainabilitySaliency2020}. 
An evaluation strategy for XAI methods proposed by \citet{agarwalOpenXAITowardsTransparentEvaluation2022} leverages a scheme to generate synthetic data, where each class is represented by a unique spatial cluster in feature space, prompting options for quantitative evaluations.
Utilizing a Visual Question-Answering (VQA) task, \cite{arrasCLEVRXAIBenchmark2022} introduce a framework, based on synthetic data, to quantify an explanation's quality for a distinctive object of an image.
Further, \cite{lesion_xai} used XAI methods to find structural changes of the ageing brain, which allowed the authors to identify white matter lesions associated to the ageing brain and forms of dementia. 
They applied layerwise relevance propagation \citep[LRP,][]{bachPixelWiseExplanationsNonLinear2015} and compared the resulting heat maps with white matter lesion maps.
\citet{ChertiEffectPretraining2021} analysed the effect of pre-training on model transfer in medical imaging but did not investigate aspects of explainability.}
 \textcolor{black}{Finally, our own work has highlighted common misinterpretations of XAI methods. Using counterexamples and analytical derivations, \citet{haufeInterpretationWeightVectors2014} and \citet{wilming2023theoretical} demonstrated that many popular XAI methods systematically attribute importance to so-called supressor variables \citep{congerRevisedDefinitionSuppressor1974, friedmanGraphicalViewsSuppression2005}, which are beneficial to the model's performance due to statistical correlations with other, informative features, but are themselves statistically unrelated to the predicted target variable. This undesirable effect was shown to be present even for linear models often assumed to be `intrinsically interpretable' \citep{rudinStopExplainingBlack2019}, and is incentivized by current algorithmic operationalizations of explanation correctness such as faithfulness \citep{bachPixelWiseExplanationsNonLinear2015}. We further devised low-dimensional benchmarks to study this effect and compare different model architectures and XAI methods using a theoretically well-founded data-driven definition of explanation correctness \citep{wilmingScrutinizingXAIUsing2022,clark2023xai}.
\citet{wilmingScrutinizingXAIUsing2022}, for example, introduce a synthetic data generation process explicitly defining discriminative and class-agnostic features such as suppressor variables. However, as these works are based on low-dimensional mathematical toy problems, the emergence of suppressor variables in realistic medical use cases such as MRI classification, and their potential influence on explanation performance in such a context, has not been studied. The present work aims to fill this gap by providing a relatively realistic clinical MR image generation scheme that induces, to a certain extent, the emergence of suppressor variables. Our work thereby provides a more challenging setting than comparable studies involving ground-truth data in medical and non-medical contexts.}

\section{Methods}

\begin{figure}
    \centering
    \includegraphics[width=0.9\linewidth]{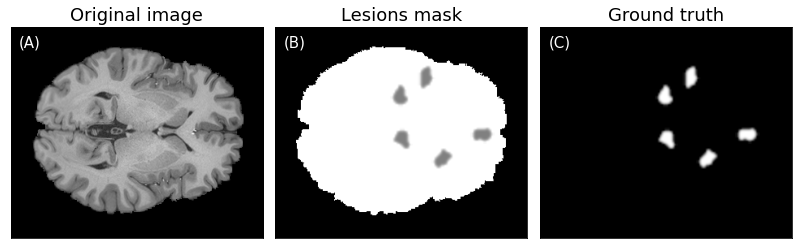}
    \caption{Example of the lesion creation process. (A) depicts an original axial MRI slice from the Human Connectome Project and denoted as background image $B$. (B) showcases an example lesion mask $L$ composed of several lesions. (C) represents ground-truth explanations used for the XAI performance study.}
    \label{fig:lesion-creation} 
\end{figure}

\subsection{Data Generation}\label{sec:data-generation}
To generate the data used for this analysis, we use 2-dimensional T1-weighted axial MRI slices from 1007 healthy adults aged between 22 and 37 years, sourced from the Human Connectome Project \cite[HCP,][see also Supplementary Material, Section~1]{hcp}.%\ref{app:sec:hcp}).
The MRI data consists of 3D MRI slices pre-processed with the FSL \citep{FSL} and FreeSurfer \citep{FreeSurfer} tools as described in \cite{hcp_preprocess1,hcp_preprocess2} and defaced as reported in \citet{defacing}. These slices provide the background on which a random number of artificial `lesions' are overlaid. Regular and irregular lesions are generated and added to the slices. Each slice contains only one type of lesions. This defines a binary classification problem, which we solve using CNNs. The lesions are added so that the dataset created is balanced.

For this study, we keep only slices with less than 55\% black pixels.
These images are 260$\times$311 pixels in size. To obtain square images, they are padded vertically with zeros and cropped horizontally. 
The final size of the slices result in background images $B \in \mathbb{R}^{270\times 270}$, where we keep the intensity values $B_{ij} \in [0, 0.7]$ for $i,j=1, \dots, 270$. 

Artificial lesions are created from a 256$\times$256 pixel noise image, to which a Gaussian filter with a radius of 2 pixels is applied.
The Otsu method \citep{Otsu} is used to binarise the smoothed image.
After the application of the morphological operations erosion and opening, a second erosion is applied to create more irregular shapes (see supplemental material section 3). %\ref{app:sec:lesion-generation}).
Since these shapes occur less frequently than regular shapes, these determine the number of different noise images necessary to create a given number of lesions.

From the images obtained after the application of the morphological operations, the connected components (contiguous groups of non-zero intensity pixels fully surrounded by zero intensity pixels) are identified, which serve as lesion candidates.
Further, lesions are selected based on the compactness of their shape.
Here, it is sufficient to consider the isoperimetric inequality on a plane $A \leq p^2 / 4 \pi$, where $A\in \mathbb{R}$ is the area of a particular lesion shape and $p \in \mathbb{R}$ its perimeter. 
The compactness is obtained by comparing the shape of the lesion candidate to a circle with the same perimeter.
The larger the compactness, the rounder the shape. 
Here, regular lesions are required to have a compactness above $0.8$ and irregular lesions have a compactness below $0.4$. 
After selecting the lesions, they are padded with a 2-pixel margin, and a Gaussian filter with a radius of $0.75$ pixels is applied to smooth the lesion boundaries. 
Examples of obtained lesions are displayed in Figure \ref{fig:example_images}. 

Three to five lesions of the same type (regular or irregular) are composed in one image $L \in \mathbb{R}^{270\times 270}$ in random locations within the brain, without overlapping and pixel-wise multiplied with the background MRI $B$ (see Figure \ref{fig:lesion-creation}).
For the lesions we consider the intensity values $L_{ij} \in [0,w]$, where $i,j$ correspond to pixels representing lesions. 
The parameter $w$ is a constant that controls the SNR. 
Higher $w$ values lead to whiter lesions and higher SNR, leading to easier classification and explanation tasks. In this study, we set $w=0.5$. 
Note also, that this setup may lead to the emergence of so-called suppressor variables.
These would be pixels of the background outside any lesion, which could still provide a model with information on how to remove background content from lesion areas in order to improve the model's predictions. Suppressor variables have been shown to be often misinterpreted for important class-dependent features by XAI methods \citep{haufeInterpretationWeightVectors2014,wilmingScrutinizingXAIUsing2022,wilming2023theoretical}.

In parallel to the generation of the actual synthetic MR images, the same lesions are added to a black image to create ground-truth masks. We summarize the ground-truth explanations via the set 
\begin{equation}
    \label{eq:lesion-ground-truth}
    \mathcal{F}_{lesions} \coloneqq \{i,j \in [270] \mid L_{ij} \neq 1 \, \text{or} \, L_{ij} \neq 0\} \; ,
\end{equation}
where $[270] \coloneqq \{1, \dots, 270\}$. The ground-truth explanation $\mathcal{F}_{lesions}$ is different for each image and an example of $\mathcal{F}_{lesions}$ represented as an image can be seen in Figure~\ref{fig:lesion-creation} (C).

Out of the $1\:006$ subjects in the HCP dataset, 60\% were used to create the training dataset, 20\% to create the validation dataset, and another 20\% to create the holdout dataset, corresponding to $24\:924$, $8\:319$, and $8\:319$ slices, respectively.
% The bottom row shows the same images with the ground-truth highlighted. 

\subsection{Pre-training}
We apply the XAI methods to the VGG-16 \citep{simonyanVGG2015} architecture, included in the Torchvision package, version 0.12.0+cu102.
Two models are pre-trained using two different corpora, and serve as starting points for our study. The first model is pre-trained using the ImageNet dataset \citep{imagenet} (out-of-domain pre-training). The weights used are included in the same version of Torchvision. The second model is pre-trained using MRI slices extracted from the HCP as described before but without artificial lesions (within-domain pre-training). Here, the task is to classify slices according to whether they were acquired from female or male subjects.
To train the latter model, $24\:924$ slices are used, 46\% of which belong to male subjects and 54\% to female subjects. These slices are arranged into batches of 32 data points. The model is trained using stochastic gradient decent (SGD) with a learning rate (LR) of 0.02 and momentum of 0.5. The learning rate is reduced by 10\% every 5 epochs. Cross-entropy is used as the loss function. 

\subsection{Fine-tuning}
After pre-training, the models are fine-tuned layer-wise on the lesion-classification problem, with images chosen from the holdout dataset, which we split into train/validation/test again (see Supplementary Material, Section~4).% \ref{app:sec:model-training}).
Each degree of fine-tuning includes the convolutional layers between two consecutive max-pooling layers.
% , as seen in figure \ref{fig:vgg}. 
Thus, the five degrees of fine-tuning are: \textit{1conv} (fine-tuning up to the first max-pooling layer), \textit{2conv} (fine-tuning up to the second max-pooling layer), and so on, up to \textit{all} (fine-tuning of all VGG-16 layers). Weights in layers that are not to be fine-tuned are frozen. SGD and Cross-entropy loss with the same parameters as used for the pre-training are employed in this phase. However, several different LRs are used. 

\subsection{XAI methods}
We apply XAI methods from the Captum library (version 0.5.0). These methods have been proposed to provide `explanations' of the models' output in the form of a heat map $\mathbf{s} \in \mathbb{R}^{270 \times 270}$, assigning an `importance' score to each input feature of an example. 
We use the default settings from Captum for all XAI methods. 
Wherever a baseline -- a reference point to begin the computation of the explanation -- is needed, an all-zeros image is used. 
This is done for Integrated Gradients, DeepLift, and GradientSHAP.
The absolute value of the obtained importance score or heat map constitutes the basis for our visualisations and quantitative explanation analyses. 
For visualisation purposes, we  further transformed the intensity of the importance scores by $-\log (1-\mathbf{s}_{ij}(1-\nicefrac{1}{b}) )/ \log(b)$, where $\log$ is the natural logarithm and $b = 0.5$. The XAI methods used were Integrated Gradients \citep{grad}, Gradient SHAP \citep{gradshap}, LRP \citep{bachPixelWiseExplanationsNonLinear2015}s, DeepLIFT \citep{deeplift}, Saliency \citep{saliency}, Deconvolution \citep{deconv} and Guided Backpropagation \citep{backprop}.

\subsection{Explanation performance}
Our definition of quantitative explanation performance is the precision to which the generated importance or heat maps resemble the ground-truth, i.e. the location of the lesions (cf. Figure \ref{fig:lesion-creation}). 
It would be expected that the best explanation would only highlight the pixels of the ground-truth, since those are the ones that are relevant to the classification task at hand.
We determine the explanation performance by finding the $n$ most intense pixels $\mathrm{Top}_{n(\mathbf{s})}$ of the heat map $\mathbf{s}$, where $n(\mathbf{s}) \coloneqq |\mathcal{F}_{lesions}|$ is the number of pixels in the ground-truth of each image.
Then we calculate the number of these pixels that were in the ground-truth (true positives). 
The precision or explanation performance $\mathbf{EP}$ is obtained by calculating the ratio between the true positives and all positives (the number of pixels in the ground-truth)
\begin{equation}
    \label{eq:explanation-performance}
    \mathbf{EP} 
    \coloneqq \frac{|\mathrm{Top}_{n(\mathbf{s})} \cap \mathcal{F}_{lesions}|}{|\mathcal{F}_{lesions}|} \, .
\end{equation}

\subsection{Baselines}
The performance of each explanation is then compared to several baseline methods, which act as `null models' for explanation performance. 
These baselines are models that are initialised randomly and not trained (random model) and two edge detection methods, the Laplace and Sobel filters.

\section{Experiments}
Showcasing the proposed dataset's utility, we fine-tune two VGG-16 models that have been previously pre-trained with the two corpora (ImageNet and MRI), to five different degrees. 
For each degree of fine-tuning, we fine-tuned 15 models with different seeds. 
Then we select the three best-performing models, where performance is measured on test data in terms of accuracy. 
We further analyse the model explanation performance of common XAI methods with respect to the ground-truth explanations in the form of lesion maps provided by our dataset.
A reference to the Python code to reproduce our experiments is provided in the Supplementary Material, Section~2.
\section{Results}
All models, except the least fine-tuned ones (1conv), reached accuracies above 90\%. 
The models pre-trained with ImageNet achieved higher accuracy than the ones pre-trained with MR images.

\subsection{Qualitative analysis of explanations}

\begin{figure*}[h!]
    \centering
    \includegraphics[width=\textwidth]{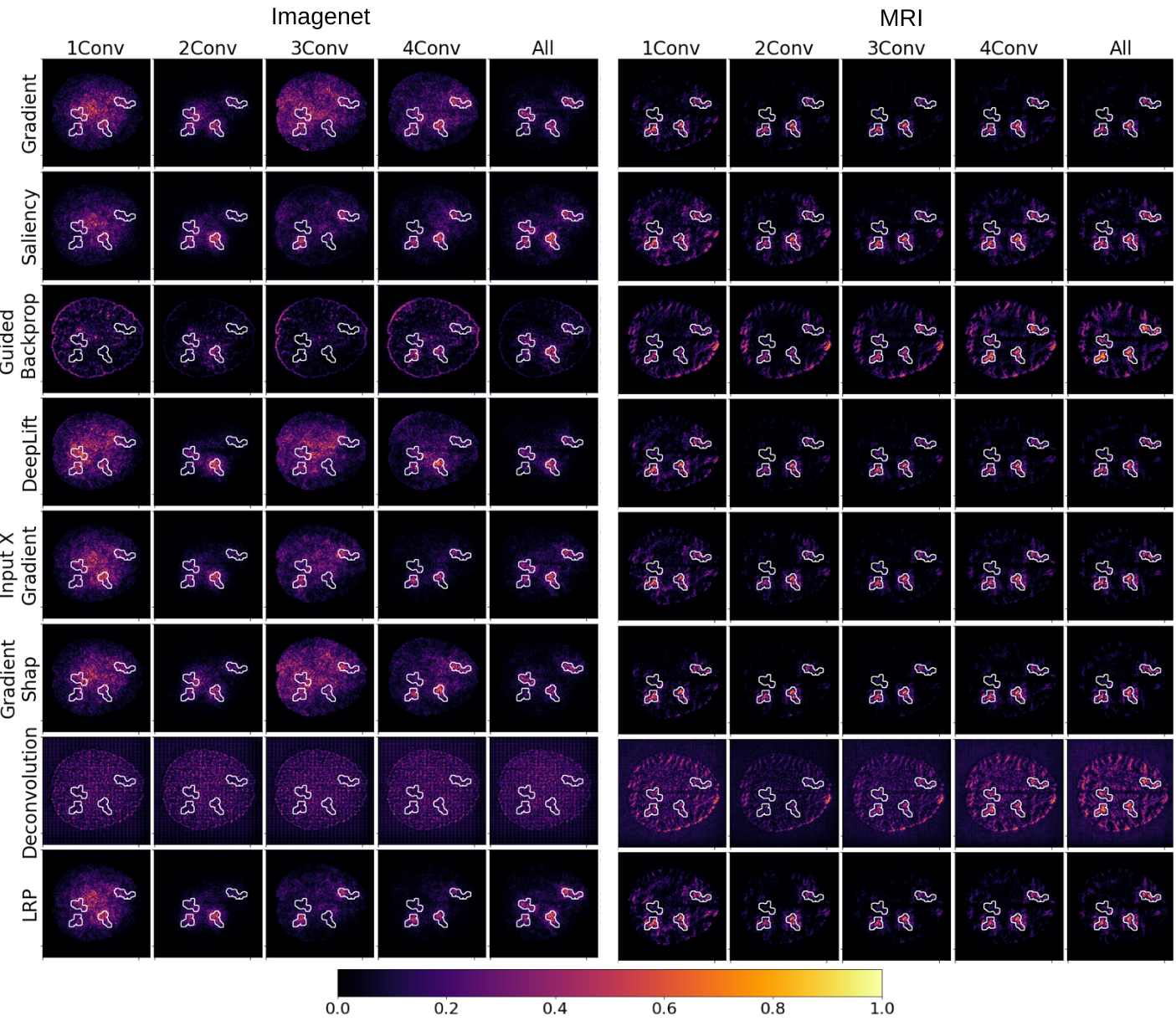}
    \caption{Examples of heat maps representing importance scores attributed to individual inputs by popular XAI methods for several degrees of fine-tuning of the VGG-16 architecture. The models were selected to achieve maximal test validation accuracy. Each row corresponds to an XAI method, whereas each column corresponds to a different degree of fine-tuning from 1 convolution block (1conv) to the entire network (all). The image is divided into two vertical blocks, where importance maps obtained from models pre-trained with ImageNet data are depicted on the left, and importance maps obtained from models pre-trained with MR images are depicted on the right.}
    \label{fig:heatmapbest}
\end{figure*}

Figure~\ref{fig:heatmapbest} displays importance heat maps for a test sample with four irregular lesions. These explanations are obtained by eight XAI methods for five degrees of fine-tuning. Plots are divided into two sections reflecting the two corpora used for pre-training (ImageNet and MRI female vs. male). The white contours in each heat map represent the ground-truth of the explanation. A good explanation should give high attribution to regions inside the white contour and low everywhere else. In this respect, most of the explanations appear to perform well, identifying most of the lesions, especially for high degrees of fine-tuning. However, the explanations generally do not highlight all of the lesions in the ground-truth. 
This image also shows that, for some XAI methods, the explanation may deteriorate for an intermediate degree of fine-tuning, and then improve again. This can be seen especially in the results of the model pre-trained with ImageNet data.
Heat maps of the untrained baseline model are shown in Section~6 of the Supplementary Material.

When comparing the ‘explanations’ obtained from models pre-trained on ImageNet data with the ones from models pre-trained on MRI data, the latter seems to contain less contamination from the structural features of the MRI background, especially for Deconvolution and Guided Backpropagation. We can further argue that some models seem to do a better job identifying the lesions than others. Particularly noisy explanations are obtained with Deconvolution, especially for models pre-trained with ImageNet data. In this case, pixels with higher importance attribution seem to form a regular grid, roughly covering the shape of the brain of the underlying MRI slice. For models pre-trained on the MRI corpus, Deconvolution is able to place higher importance within the lesions for higher degrees of fine-tuning.

\subsection{Quantitative analysis of explanation performance}

\begin{figure*}[h!]
    \centering\includegraphics[width=\textwidth]{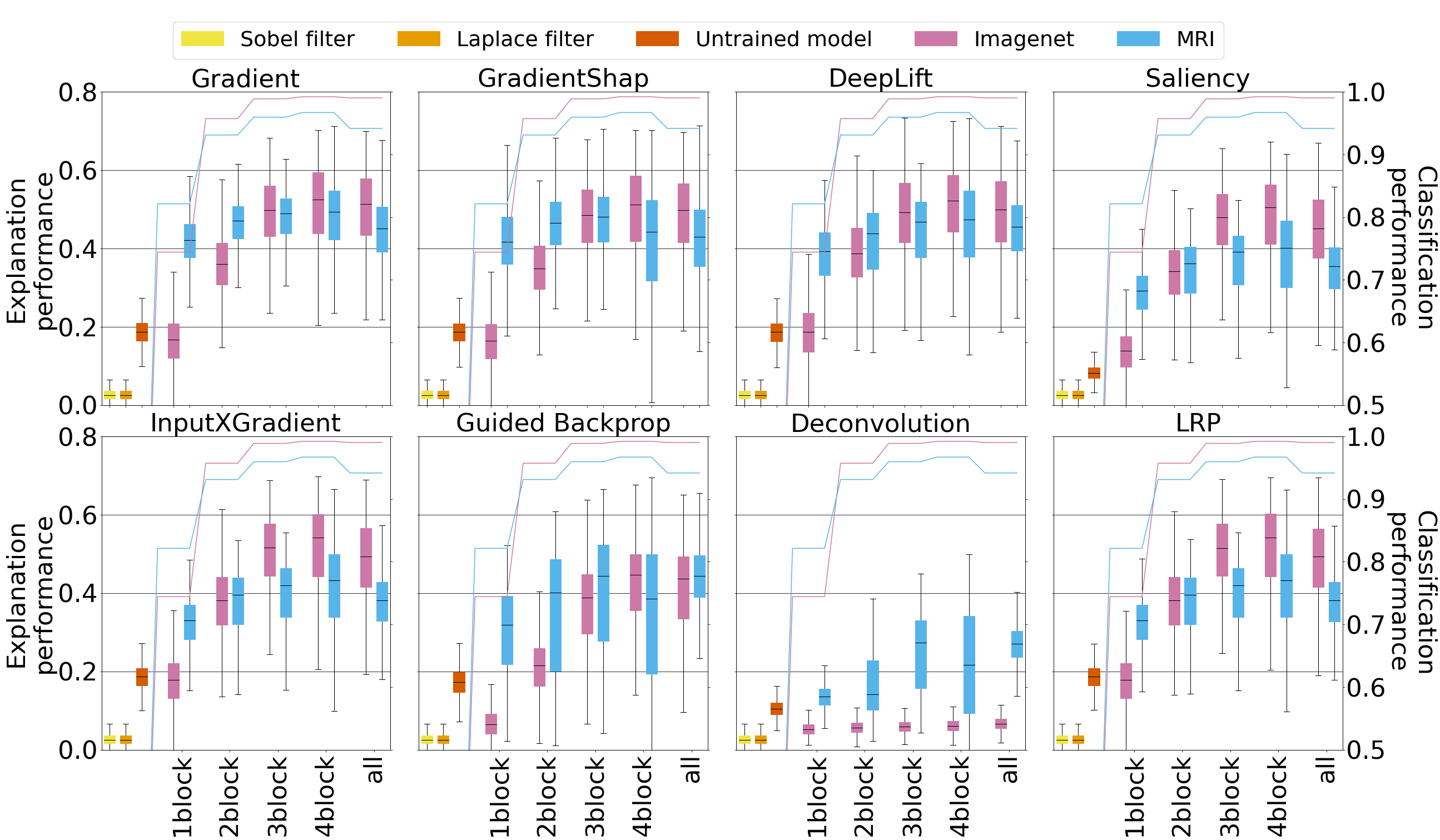}    \caption{Quantitative explanation performance for XAI methods applied to the five best models, with different degrees of fine-tuning. The blue line and boxplots correspond, respectively, to the classification performance (accuracy) and explanation performance (precision) derived from models pre-trained with MRI data, whereas the pink line and boxlots correspond analogously to classification and explanation performance for models pre-trained with ImageNet data. The other three boxplots correspond to the performance of baseline heat maps. Yellow and orange colour correspond to the Sobel and Laplace filters respectively, and red colour to the model with random weights.}
    \label{fig:boxplots_best}
\end{figure*}

Figure~\ref{fig:boxplots_best} shows quantitative explanation performance.
Here, each boxplot was derived from the intersection of test images that were correctly classified by all models ($N=2\:371$). 
The results obtained for the edge filter baseline as well as the random baseline model are derived from the same $2\:371$ images. 
Note that the edge detection filters only depend on the given image and are independent of models and XAI methods. 
Thus, identical results are presented for edge filters in each subfigure. 
The lines in the background correspond to the average classification performance (test accuracy) of the five models for each degree of fine-tuning.
The random baseline model is only one and has a test accuracy of $50\%$. 
Interestingly, models pre-trained with ImageNet data consistently achieved higher classification performance than models pre-trained with MR images. 
The classification performance of the models pre-trained with MR images peaks at an intermediate degree of fine-tuning (3conv), while the models pre-trained with ImageNet improve with higher fine-tuning degrees.

In some settings, ImageNet pre-training leads to considerably worse explanation performance. This is the case for specific methods such as Deconvolution and, to some extent, Guided Backpropagation. Moreover ImageNet pre-training leads to worse explanations across all XAI methods for lower degrees of fine-tuning (1conv and 2conv), where large parts of the models are prohibited to depart from the internal representations learned on the ImageNet data.

As a function of the amount of fine-tuning, explanation performance generally increases with higher degrees of fine-tuning. However, depending on the XAI method used, and the corpus used for pre-training, this trend plateaus or even slightly reverses at a high degree of fine-tuning (4conv). 

Importantly, explanation performance appears to strongly correlate with the classification performance of the underlying model.
As classification accuracy could represent a potential confound to our analysis, we repeated our quantitative analysis of explanation performance based on five models with similar classification performance per pre-training corpus and degree of fine-tuning. 
Here, it is apparent that, when controlling for classification performance, models pre-trained on MRI data consistently outperform equally well-predicting models that were pre-trained on ImageNet data in terms of explanation performance.
These results are presented in \textcolor{black}{Section~7 of the Supplementary Material.} 

\section{Discussion}
The field of XAI has produced a plethora of methods whose goal it is to `explain' predictions performed by deep learning and other complex models, including CNNs. However, quantitative evaluations of these methods based on ground-truth data are scarce. 
Even if these methods are based on seemingly intuitive principles, XAI can only serve its purpose if it is itself properly validated, which is so far not often done.
The present study was designed to create a benchmark within which explanation quality can be objectively quantified. 
To this end, we designed a well-defined ground-truth dataset for model explanations, where we modelled artificial data to resemble the important clinical use case of structural MR image classification for the diagnosis of brain lesions. 
With this benchmark dataset, we propose a framework to evaluate the influence of pre-training on explanation performance.

We observed a correlation between classification accuracy and explanation performance, which could be expected since a more accurate model is likely to more successfully focus on relevant input features. 

Networks trained on ImageNet data may have learned representations for objects occurring only outside the domain of brain images (e.g., cats and dogs).
The existence of such representations in the network seems to negatively affect XAI methods, whose importance maps are in parts derived by propagating network activations backwards through the network.
Consistent with this remark is the observation that for lower degrees of fine-tuning (1conv and 2conv), the explanation quality of models pre-trained with ImageNet data is worse compared to models pre-trained with MR images. 
These findings challenge the popular view that the low-level information captured by the first layers of a CNN can be shared across domains. 

Our quantitative analysis suggests a large dispersion of explanation performance for all XAI methods, which may be unexpected given the controlled setting in which these methods have been applied here. 
Individual explanations can range from very good to very poor even for high overall classification accuracy, indicating a high risk of misinterpretation for a considerable fraction of inputs.

\subsection*{Limitations}

Note, our analysis of XAI methods is limited to one DNN architecture, VGG-16, mainly showcasing the utility of our devised ground-truth dataset for model explanations.
We stress, that, rather than conducting an exhaustive study of the behaviour of popular XAI methods in relation to specific model architectures, with our work, we aim to predominantly contribute to the evaluation of XAI methods by providing a controlled ground-truth dataset, with known explanations, class-related features, enabling future research to benchmark new XAI methods.

We emphasize that we purposely refrain from expert annotated data as it does not constitute a stable ground-truth, and that full knowledge about the underlying ground-truth is needed to validate methods, a purpose that is only served be synthetically crafted data. In the medical domain, ML methods are often used with the expectation that they will uncover statistical relations that are either unknown or too complex (e.g. involving non-linear interactions of features) for human experts to discern.
When experts annotate data, they may inadvertently overlook these features, potentially leading to false-positive detections if an XAI method indeed succeeds in highlighting them. Conversely, human experts
may provide annotations that are simply incorrect. 
They can be influenced by pseudo-correlations in the data resulting from limited sample size in prior studies, or mistakenly base their judgement on confounders or even suppressors. 
In such instances, a correctly functioning XAI method may be mistakenly accused of delivering false-negative detections.
Note in this context that clinical doctrines are highly fluctuating as
new evidence is constantly being produced. For example, the assumed causal role of beta-amyloid and tau protein plaques in the brain for various types of dementia is currently being challenged.
To address these challenges and strive towards real-world validity, experiments involving annotated real data are valuable and complementary next steps.
However, they cannot entirely replace ground-truth experiments involving synthetic or manipulated real data due to their intrinsic biases. 
And generating realistic artificial and controllable image data for the MRI domain is, in itself, a very hard problem.

Furthermore, the proposed lesion generation process resembles the idea of white matter hyperintensities where we aim to approximate specific neurodegenerative disorders from a `model perspective', where a natural prediction task would be `healthy' vs. `lesioned brain'.
But it would be difficult to define a ground-truth for the class `healthy'.
Hence, we chose to create a classification problem based on two different shapes of lesions: round vs elongated.
Admittedly, this distinction has no immediate physiological basis and serves purely the purpose of this benchmark, i.e., we can solve a classification task well enough by using a model architecture considered popular in this field.
However, we provide a classification scenario where the background, real brain slice images, provides features that are partially leveraged by ML models, which put XAI methods in the position to differentiate between class-related features, artificial lesions, and realistic brain-related features. 
Where we think that this distinction constitutes a realistic environment for XAI methods. 
In this light, our dataset can be seen as a first instance of contributing to the performance quantification of explanations produced by XAI methods for the MRI domain.

We argue that the quantitative validation of the \emph{correctness} of XAI methods is still a greatly under-investigated topic given how popular some of the methods have become.
Major efforts both on the theoretical and empirical side are needed to create a framework within which evidence for the correctness of such methods can be provided. 
As a first step towards such a goal, meaningful definitions of what actually constitutes a correct explanation need to be devised. 
While in our study, ground-truth explanations were defined through a data generation process, other definitions, depending on the intended use of the XAI, are conceivable. 
The existence of such definitions would then pave the way for a theoretical analysis of XAI methods as well as for use-case-dependent empirical validations.

\section{Conclusion}
In this work we created a versatile synthetic image dataset that allows us to quantitatively study the classification and explanation performances of CNN and similar complex ML methods in a highly controlled yet realistic setting, resembling a clinical diagnosis/anomaly detection task based on medical imaging data. Concretely, we overlaid structural brain MRI data with synthetic lesions representing clinically relevant white matter hyperintensities. We propose this dataset, to evaluate the explanations obtained from pre-trained models.

Our study is set apart from the majority of work on XAI in that it uses a well-defined ground-truth for explanations, which allows us to quantitatively evaluate the `explanation' performance of several XAI methods.

Our study revealed a strong correlation between the classification performance of the model and the explanation performance of the XAI methods. 
Despite this correlation, models fine-tuned to a greater extent were shown to lead to better explanations. 
Controlling for classification performance, models pre-trained on MRI data lead to better explanations for every XAI method. 
The explanation performance of models pre-trained on within-domain images seem to have more stable explanation performance for a bigger range of classification accuracies.
On the other hand, the explanation performance of models pre-trained with more general images quickly degrades with lower classification performance.

The quantitative analysis of the explanations also shows a concerning variability of explanation performance values, suggesting that, when these methods are used to explain an individual prediction, a large uncertainty is associated with the correctness of the resulting importance map. This is a critical issue when using XAI methods to `explain' predictions in high-stake fields such as medicine. 

\section*{Acknowledgements}
This result is part of a project that has received funding from the European Research Council (ERC) under the European Union’s Horizon 2020 research and innovation programme (Grant agreement No. 758985), the German Federal Ministry for Economy and Climate Action (BMWK) in the frame of the QI-Digital Initiative, and the Heidenhain Foundation.

\bibliographystyle{icml2025}
\bibliography{references}

\end{document}